\documentclass[runningheads]{llncs}
\usepackage{graphicx}
\usepackage{pifont}
\usepackage{pifont}%
\newcommand{\cmark}{\color{green}\ding{51}}%
\newcommand{\xmark}{\color{red}\ding{55}}%

\usepackage{tikz}
\usepackage{comment}
\usepackage{amsmath,amssymb} %
\usepackage{booktabs}
\usepackage{multirow}
\usepackage[pagebackref,breaklinks,colorlinks]{hyperref}

\usepackage[accsupp]{axessibility}  %

\usepackage{xcolor,colortbl}
\newcommand{\bgl}{\cellcolor[HTML]{DDDDDD}}
\newcommand{\bgd}{\cellcolor[HTML]{BBBBBB}}

\usepackage[capitalize]{cleveref}
\usepackage{siunitx}
\usepackage{mathtools}

\newcommand{\norm}[1]{\left\lVert#1\right\rVert}

\newcommand{\R}{\mathbb{R}}
\newcommand{\SE}{\ensuremath{\mathbf{SE}}}

\newcommand{\etal}{\textit{et al}. }

\newcommand{\plucker}{Plücker~}

\begin{document}
\pagestyle{headings}
\mainmatter
\def\ECCVSubNumber{2290}  %

\title{Generalizable Patch-Based Neural Rendering} %

\titlerunning{Generalizable Patch-Based Neural Rendering}
\author{Mohammed Suhail\inst{1} \and
Carlos Esteves\inst{4} \and
Leonid Sigal\inst{1,2,3}\and 
Ameesh Makadia\inst{4}
}
\authorrunning{M. Suhail et al.}
\institute{University of British Columbia \email{\{suhail33,lsigal\}@cs.ubc.ca} \and Vector Institute for AI \and
Canada CIFAR AI Chair \and
Google
\email{\{machc,makadia\}@google.com}
}

\maketitle

\begin{abstract}
  Neural rendering has received tremendous attention since the advent of Neural Radiance Fields (NeRF), and has pushed the state-of-the-art on novel-view synthesis considerably. The recent focus has been on models that overfit to a single scene, and the few attempts to learn models that can synthesize novel views of unseen scenes mostly consist of combining deep convolutional features with a NeRF-like model.
  We propose a different paradigm,
  where no deep visual features and no NeRF-like volume rendering are needed.
  Our method is capable of predicting the color of a target ray in a novel scene directly,
  just from a collection of patches sampled from the scene.
  We first leverage epipolar geometry to extract patches
  along the epipolar lines of each reference view.
  Each patch is linearly projected into a 1D feature vector and
  a sequence of transformers process the collection.
  For positional encoding, we parameterize rays as in a light field representation,
  with the crucial difference that the coordinates are canonicalized with respect to the
  target ray, which makes our method independent of the reference frame and improves generalization.
  We show that our approach outperforms the state-of-the-art on novel view synthesis of unseen scenes
  even when being trained with considerably less data than prior work.
  Our code is available at \url{https://mohammedsuhail.net/gen_patch_neural_rendering/}.
\end{abstract}

\begin{figure}[t!]
    \centering
    \includegraphics[width=.8\textwidth]{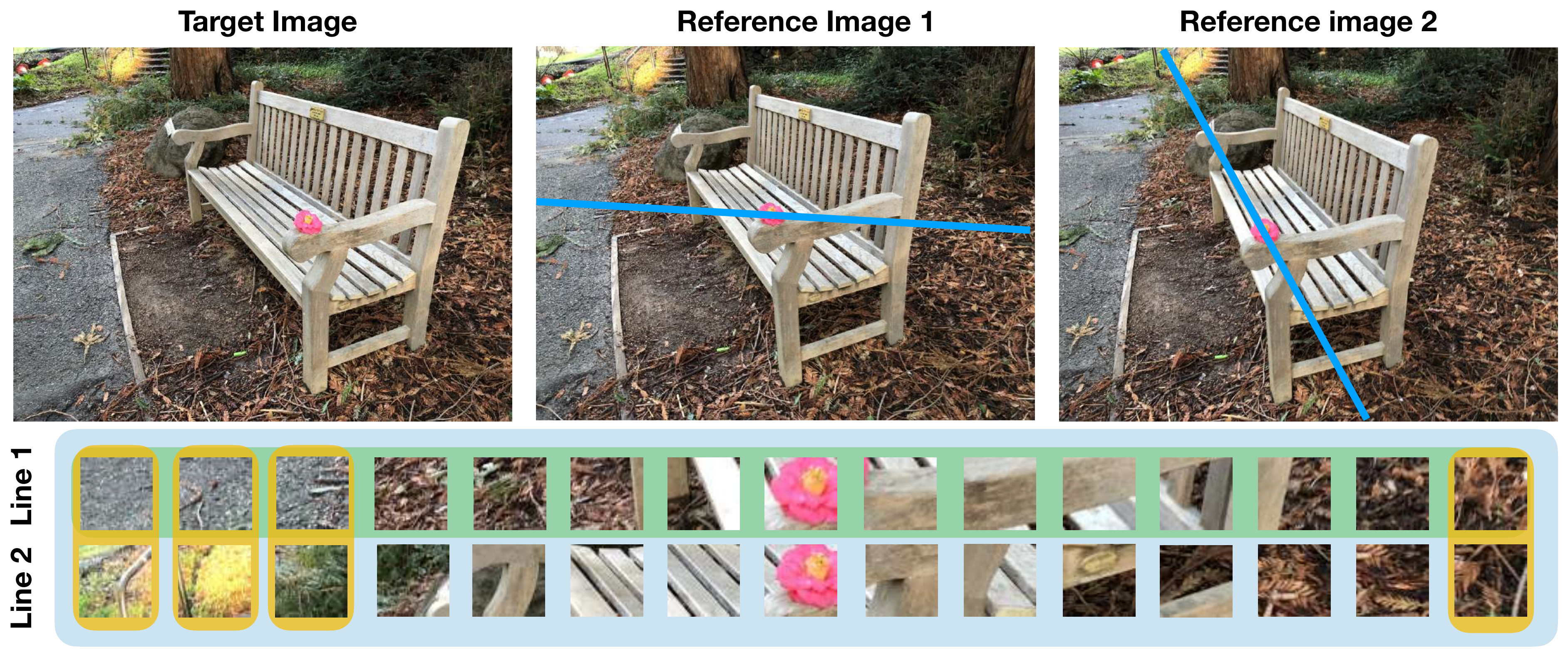}
    \caption{\textbf{Motivation overview.}
      Our goal is to predict the color of a target ray,
      given only the reference images and camera poses.
      Consider the patches along each epipolar line,
      which correspond to samples of increasing depth along the target ray.
      If there are many matching patches at some depth,
      there is a high chance that the patch around the target ray also matches.
      In this example, the matching patches contain the flower,
      which is where the target ray hits.
      This motivates our three-stage architecture,
      that first exchanges information along views at each depth (yellow),
      then aggregates information along depths for each view (green),
      and finally aggregates information among reference views to predict the ray color (blue).
      The figure shows only 2 reference views with 15 sampled patches each,
      but in practice we use a larger number of views and samples.
    } 
    \label{fig:patch}
  \end{figure}
  
\section{Introduction}
Synthesizing novel views of a scene from a set of images obtained from different viewpoints
is a long-standing problem in computer graphics and computer vision.
Recent advances in this problem~\cite{tewari2021advances}
employ neural networks to learn scene representations (neural scene representations),
combined with classical volume rendering to produce a novel view from any desired viewpoint,
an idea spearheaded by NeRF~\cite{mildenhall2020nerf}.
Most of these methods are trained by overfitting to a single scene in order to produce arbitrary novel views of that same scene.
While capable of producing high-quality photorealistic images,
the need for retraining on each new scene limits their practical application.

In this paper, we consider the more difficult task of training a single model
that is capable of generating novel views of unseen scenes. 
There are a few notable efforts in this direction~\cite{Chen_2021_ICCV,wang2021ibrnet,Yu_2021_CVPR}.
One key idea of these methods is to augment NeRF inputs with deep convolutional features,
which include both local and global context.
However, these methods still rely on scene-specific inputs such as 3D positions and directions,
which are not reliable on unseen scenes.
We also hypothesize that using feature extractors 
that have large receptive fields such as UNet~\cite{ronneberger2015u} 
or Feature Pyramid Networks~\cite{lin2017feature} 
is harmful when generalizing to scenes visually far from the training distribution. 

We propose a different approach that takes only local linear patch embeddings as input,
eschewing deep convolutional networks.
Moreover, our method does not require the ubiquitous volume rendering from NeRF; it produces the color of a target pixel directly from a set of reference view patches. 

We are inspired by both classical and recent works.
Classical computer vision tasks such as optical flow and image feature matching for 3D reconstruction
were historically dominated by techniques operating on local patches~\cite{Harris88acombined,sift,lucaskanade,shitomasi}.
In fact, for some tasks the classical methods still outperform modern deep learning ones~\cite{Schonberger_Comparative_Evaluation_CVPR_2017}.
Another example is COLMAP~\cite{schoenberger2016sfm,schoenberger2016mvs}
which is a widely popular method for 3D reconstruction and typically used to generate camera poses
(and sometimes depth maps) that are inputs to modern neural rendering.

Our decision to focus on local patches
and to avoid convolutional features is supported by the recent success of the Vision Transformers
(ViT)~\cite{dosovitskiy2020image}, which we employ.
A second reason to use transformers~\cite{vaswani2017attention} is that our input
is effectively a set of patches, and self-attention is a powerful mechanism to learn from sets
without making any assumption about the order of the elements.
We show that transformers can effectively replace both the convolutional features and
the volume rendering typically employed in the tasks we consider.

Our key contribution is to leverage the structure of the patch collection
to build a multiview representation
that is further refined along epipolar lines and reference views to
predict the final color.
\Cref{fig:patch} explains the idea.
Another unique aspect of our method is the canonicalized positional encoding
of rays, depths, and camera poses, which is independent of the frame of reference,
enabling superior generalization performance. 

\medskip
\noindent
{\bf Contributions:} Our contributions can be summarized as follows,
\begin{itemize}
\item We introduce a model that renders target rays in unseen scenes
  directly from a collection of patches sampled along epipolar lines of reference views.
\item To exploit the structure of the patch collection, we design an architecture with
  stacked transformers operating over different subsets of the collection such that
  features are learned, combined, and aggregated in principled ways. 
\item To improve generalization to unseen scenes, we introduce canonicalized positional encodings
  of rays, depths, and camera poses such that
  all inputs to the model are independent of the scene's frame of reference. 
\item Our model outperforms previous baselines in multiple train and evaluation datasets,
  while using as little as 11\% of training data in certain cases.
  
\end{itemize}

\section{Related Work}

\subsection{Neural scene representations}
Our method is in the broad category known as neural rendering, where
neural networks are used to represent a scene and/or directly render
views~\cite{tewari2021advances}. Neural fields~\cite{xie2021neuralfield} are also closely related. 
The majority of recent methods employ neural scene representations coupled with classical rendering methods, as popularized by NeRF~\cite{mildenhall2020nerf}.
These works can be broadly classified into models that represent the scene using a {\em surface} or {\em volumetric} representation~\cite{tewari2021advances}. Surface representation methods either explicitly represent the scene as point clouds \cite{aliev2020neural,lassner2021pulsar,pfister2000surfels,ruckert2021adop,Wiles_2020_CVPR,yifan2019differentiable}, meshes \cite{burov2021dynamic,hu2021worldsheet,thies2019deferred}, or implicitly using signed distance function \cite{chen2019learning,kellnhofer2021neural,park2019deepsdf,takikawa2021neural,yenamandra2021i3dmm}. Volumetric representations on the other hand typically use voxel grids \cite{nguyen2019hologan,sitzmann2019deepvoxels}, octrees \cite{liu2020neural,yu2021plenoctrees}, multi-plane \cite{wizadwongsa2021nex,zhou2018stereo}, implicitly using a neural network \cite{genova2019learning,liu2020dist,niemeyer2020differentiable} or a coordinate-based network as in NeRF~\cite{mildenhall2020nerf} and its variants \cite{barron2021mip,mildenhall2021nerf,muller2022instant}. Recently, works such as Vol-SDF \cite{yariv2021volume}, NeuS \cite{wang2021neus} and UNISURF \cite{oechsle2021unisurf} propose to use volumetric rendering methods to extract a surface representation.

Our method differs from these because there is no structured neural scene representation
and no volume rendering -- the target pixel color is obtained directly by learning weights to
blend reference pixels, taking only a set of patches around the reference pixels as input. 
Thus, our approach fits into the category of image-based rendering.

Moreover, our model can be trained once on a set of scenes and applied to novel scenes,
which can be more efficient than re-training scene-specific models for every new scene as is common.
Concurrent work has achieved impressive results on accelerating NeRFs~\cite{yu21_plenox,mueller22_instan_neural_graph_primit_with},
providing a reasonable alternative when efficiency is important
and re-training for every scene is not a hindrance.

\subsection{Image-based rendering}
Image-based rendering methods~\cite{shum2000review,shum2008image} typically construct novel views of a scene by warping and compositing a set of reference images.
Shum and Kang~\cite{shum2000review} classified most of these works into categories
that use {\em no} geometry, {\em explicit} geometry or {\em implicit} geometry.
Methods that do not model the geometry rely on the characterization of the plenoptic function.
Light field rendering~\cite{levoy1996light} is one such method that used 4D light field plenoptic function to render novel views by interpolating a set of input samples.
Light field rendering, however, requires a dense sampling of input views to be accurate.
Follow-up works such as Lumigraph~\cite{gortler1996lumigraph} incorporate approximate geometry to overcome the dense sampling requirement. 
Explicit geometry based methods \cite{hedman2017casual,hedman2018instant,riegler2020free,riegler2021stable} generate a geometric reconstruction of the scene in the form of a 3D mesh.
However, explicit 3D reconstruction without 3D supervision is a hard learning problem,
and undesirable artifacts in the reconstructed geometry impact rendering quality.
Implicit geometry methods \cite{viewinterpolation,viewmorphing}
rely on aggregating multiple input views to synthesize a novel view.
Recently, LFNR~\cite{suhail21_light_field_neural_render} proposed to use epipolar geometry in conjunction with light field ray representations to model view-dependent effects. Other works~\cite{attal2022learning,feng2021signet,sitzmann2021light} similarly have explored neural representations for light field rendering.
Part of our architecture is similar to LFNR; however,
our method is aimed at generalizing to unseen scenes as opposed to overfitting on a single scene,
which avoids expensive retraining for each new scene.

Stereo Radiance Fields~\cite{Chibane_2021_CVPR} has a focus on efficiency
and was one of the first methods tackling generalization to novel scenes.
PixelNeRF~\cite{Yu_2021_CVPR} conditions a NeRF~\cite{mildenhall2020nerf}
on deep convolutional visual features of the reference views,
enabling generalization to new scenes; however,
it uses absolute positions and directions as inputs to the NeRF,
which generalize poorly across scenes.
Similarly, IBRNet~\cite{wang2021ibrnet} also uses deep features and NeRF-like volume rendering,
but it learns to blend colors from neighboring views for each point along a ray.
IBRNet uses the difference between view directions as MLP inputs;
while this is superior to absolute coordinates,
the relative view directions still depend on a global reference frame
which is scene-specific. 
MVSNeRF~\cite{Chen_2021_ICCV} constructs a cost volume from deep visual features.
The voxel features are then concatenated to the usual NeRF inputs
including absolute positions and directions for rendering novel views.

In contrast with these works, our method
1) does not require deep convolutional features,
operating directly on linear projections of local patches, similarly to ViT~\cite{dosovitskiy2020image};
2) does not require volume rendering,
producing the final colors directly from a reference set of patches;
and 3) is independent of the input frame of reference,
leveraging canonicalized ray, point and camera representations,
which improves its generalization ability.
Concurrent work on neural rendering generalizable to unseen scenes include
GeoNeRF~\cite{johari2021geonerf} and NeuRay~\cite{liu2021neural},
but both require at least partial depth maps during training.

\subsection{Transformers in vision}
Popularized by Vaswani \etal~\cite{vaswani2017attention}, transformers are sequence-to-sequence models that use an attention mechanism to incorporate contextual information from relevant parts of the input.
Initially developed for NLP tasks~\cite{chernyavskiy2021transformers}, transformer-based models have also achieved
state-of-the-art on a variety of vision problems~\cite{dosovitskiy2020image,carion2020end,lu2019vilbert,chang2022maskgit,suhail21_light_field_neural_render}.

Recently, Robin~\etal~\cite{rombach2021geometry} proposed the use of transformers to generate novel views from a single image without explicit geometric modeling.
Scene representation transformers~\cite{sajjadi2021scene} similarly presented a model for novel view synthesis using self-supervision from images. Their experiments, however, are limited to low resolution images (maximum size of $178 \times 128$ pixels).
Slightly more related to our approach are IBRNet~\cite{wang2021ibrnet},
which employs a ray-transformer module
to estimate densities via self-attention over samples along the ray,
and NerFormer~\cite{reizenstein21co3d}, which alternates self-attention over views and rays,
but is object-based and aims to generalize only to new instances of the same object category. 

Our use of transformers differs greatly from such works because
(i) we use transformers in all stages, from the patch embedding to final target ray color prediction,
not requiring deep convolutional features nor volume rendering, and
(ii) we design a unique architecture with three different transformers operating along and collapsing
different dimensions.

\section{Approach}
\begin{figure}[t!]
    \centering
    \includegraphics[width=\textwidth]{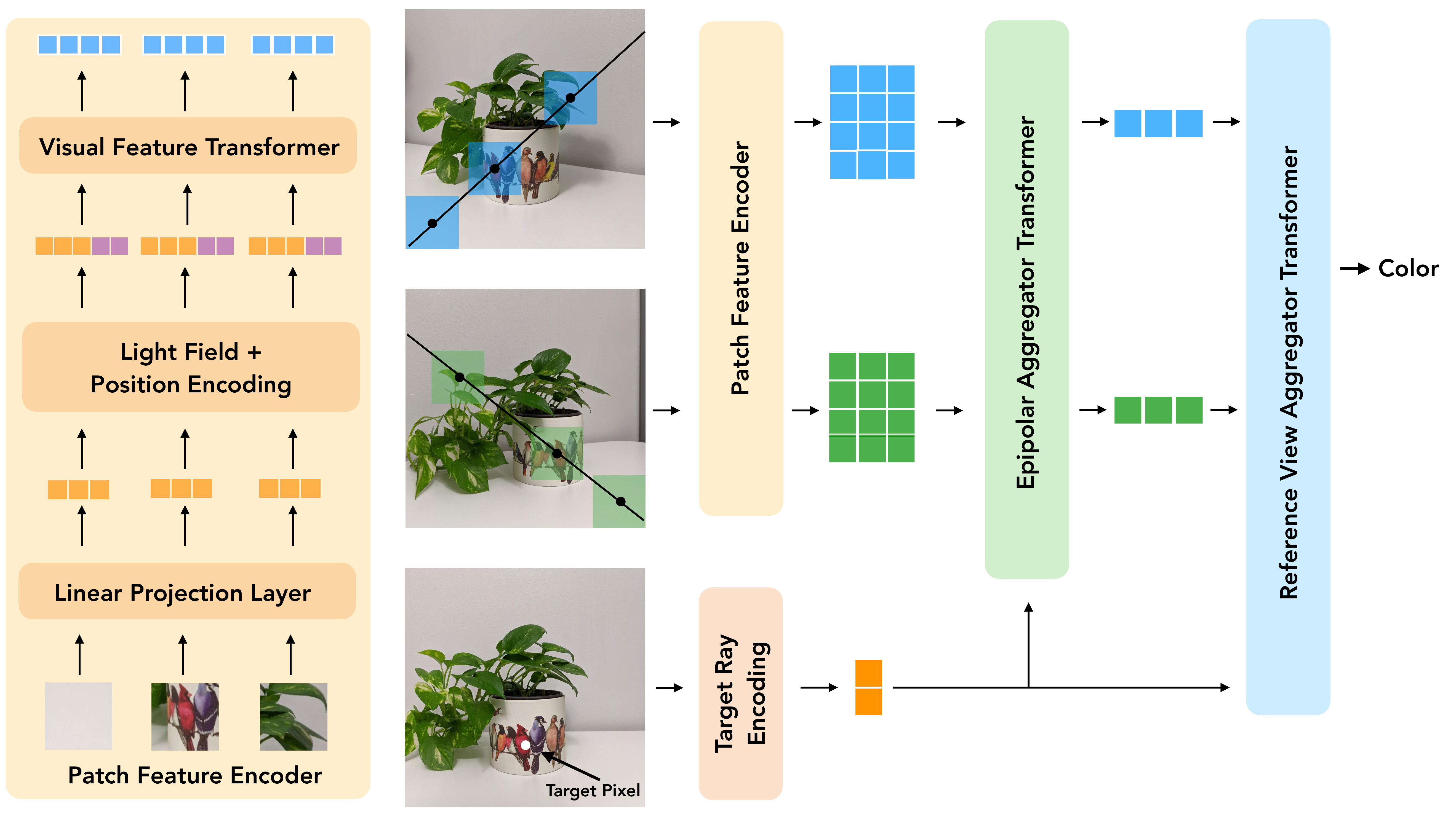}
    \caption{\textbf{Model Overview.}
      Our model consists of three stages, with a different transformer per stage.
      First, patches along epipolar lines are extracted, linearly projected,
      and arranged in a grid of $K$ reference views by $M$ sampled depths.
      The first transformer takes a sequence of views
      and is repeated for each depth, returning another $K \times M$ grid.
      The second transformer takes a sequence of depths and is repeated for each view;
      it collapses features along the depth dimension, returning $K$ view features.
      The third transformer aggregates the $K$ view features.
      Attention weights extracted from the second and third transformers are used
      to blend colors over views and epipolar lines and make the final prediction.
      A canonicalized positional encoding of rays, depths and cameras is appended to
      the transformer inputs. 
    }
    \label{fig:model_overview}
\end{figure}

Given a set of scenes with a collection of images and their corresponding camera poses, we aim to learn a generic rendering model that is capable of rendering novel views of a scene without training on it.
At the core of our model is a reference-frame-agnostic rendering network that relies only on local patches observed from nearby reference cameras.
Figure \ref{fig:model_overview} provides a visual overview.
We present our approach in the following order:
first we introduce light field representations;
then we discuss the construction and embedding of reference patches;
and finally we detail our transformer-based rendering network that maps a target light field and reference patches to radiance.

\subsection{Light field representation}%
\label{sec:lf}
The light field characterizes the radiance through points in space.
It can be described by a five-dimensional function on $\R^3 \times S^2$,
mapping each direction through each point to its radiance.
In free space, the radiance along a ray remains constant,
thus allowing to parametrize the light field as a $4D$ function~\cite{levoy1996light}.

Depending on the camera configuration, different light field representations can be used.
For example, for a scene with forward-facing camera configuration, the rays can be parametrized by their intersections with two planes perpendicular to the forward direction,
a representation known as the light slab~\cite{levoy1996light}.
The entries of the $4D$ vector are the coordinates of the intersections on each plane's $2D$ coordinate system.
An alternative representation suitable for bounded scenes observed from all directions is known as the two-sphere~\cite{camahort1998uniformly},
and represents rays by their two intersections with a sphere bounding the scene.
Prior works such as LFNR~\cite{suhail21_light_field_neural_render} exploit the camera configuration information of the scene to decide the underlying parametrization.
They use light slab parametrization for forward-facing-scenes and two-sphere parametrization for $360^\circ$ scenes.

In this work, the ray representations are used as positional encoding in the transformers. 
Since we wish to generalize to new scenes and therefore cannot make assumptions
about the camera configurations, we use \plucker coordinates as the choice of parametrization. 
Given a ray through a point $o$ (the ray origin) with direction $v$, the \plucker coordinates can be obtained as $r = (v, o \times v)$.
The representation is six-dimensional, however it has only four degrees of freedom since it is defined up to a scale factor and the two vectors that compose it must be orthogonal.
The Light Field Networks~\cite{sitzmann2021light} use the same parametrization
but in a different context.

\subsection{Patch extraction}
\label{sec:patchextraction}
Given a target viewpoint, our method relies on eliciting ``local'' light field patches to produce the ouput images. To extract such patches, we first identify a set of reference images that serve as 2D slices of the plenoptic function observed from neighboring viewpoints. While our model is agnostic to the number of reference images, we use a subset of the available input images for patch extraction. 
Specifically, for a target camera we take a subset of the $N$ closest views. %
We randomly sub-sample $K$ views from this subset during training, and use the closest $K$ views for inference.

Given the set of reference images $\mathcal{I} = \{I_1, I_2,...,I_K\}$, the next step is to fragment them into patches.
Dosovitskiy~\etal~\cite{dosovitskiy2020image} split the entire image into fixed-size non-overlapping patches.
While this partition is useful for global reasoning (e.g.\ image classification), for view synthesis the relevant regions in the image can isolated by exploiting the epipolar geometry between views.
For a given image in the reference set $\mathcal{I}$, we compute the epipolar line corresponding to the target pixel.
We sample $M$ points along this epipolar line such that their $3D$ re-projections on the target ray are spaced linearly in depth.
We then extract square patches around each of the $M$ points,
and this process is repeated for all reference images.
The resulting reference patch set is indexed by view and depth:
$\mathcal{P} = \{P_k^m\ \mid 1 \le k \le K,\, 1 \le m \le M\}$.

\subsection{Patch embedding and positional encoding}%
\label{sec:posenc}
The inputs to the transformers are patch embeddings which we generate we generate by linearly projecting flattened input patches.
The patch features for the $m$-th sample along the epipolar line on view $k$ is denoted $p_k^m$.

Since transformers are agnostic to the position of each element in the input sequence,
typically a positional encoding is added to the features to represent the
spatial relationship between elements.
Unlike prior works \cite{dosovitskiy2020image},
since the location and source of patches do not remain the same across batches,
we cannot include a learnable embedding into the sequence.
Instead, we extract the geometric information associated with each patch and append them to the flattened patch feature vectors.

We use three forms of positional encoding:
\begin{enumerate}
\item To retain the reference patch position in space,
  we use the light field encoding of the rays emanating from the reference camera as described in \cref{sec:lf}.
  We represent the $m$-th ray along the epipolar line of view $k$ by $r_k^m$.
\item To retain the position of the patch in the sequence of patches along the epipolar line,
  we encode the distance along the target ray corresponding to the patch center using a sinusoidal positional encoding that follows NeRF~\cite{mildenhall2020nerf}.
The encoded distance for the $m$-th sample is represented by $d^m$.
\item To retain geometry between target and reference cameras,
we also append the relative camera pose as a flattened rotation matrix and a 3D translation,
which is shared among all patches associated to the same camera and denoted by $c_k$ for camera $k$.

\end{enumerate}

\subsection{Canonicalized ray representation}
\label{sec:canon}
Structure-from-Motion (SfM) methods that are used to estimate camera extrinsics
can only reconstruct scenes up to an arbitrary similarity transformation --
rotation, translation and scaling. 
Prior works~\cite{Chen_2021_ICCV,wang2021ibrnet,Yu_2021_CVPR} use such estimations
to compute scene specific coordinates such as view directions and $3D$ coordinates of points.

We hypothesize that for best generalization to unseen scenes,
the inputs to the model should be invariant to similarity transformations.
This means that model should produce the same result upon a change of reference frame
or rescaling of the input camera poses.
IBRNet~\cite{wang2021ibrnet} takes a step towards this idea by using the difference
between reference and target direction vectors instead of absolute directions,
but the difference is still a 3D vector that is not independent of the frame of reference,
and so are the 3D positions of points along the target ray.

The positional encoding of relative camera poses and distance values, as described in~\cref{sec:posenc},
are made invariant to similarities by
simply scaling the camera positions by the maximum depth of the scene output by SfM.

The encoding of rays in the light field, however, need to be canonicalized.
Our key idea is to define a local frame centered on each ray (not camera).
For a target pixel $x \in \mathbf{RP}^2$ in the target camera with extrinsics $[R \mid t]$ and
intrinsics $C$, we first obtain the corresponding ray direction $v = R^\top C^{-1}x$.
We use $v$ and the camera $y$ axis to determine the local frame.
Specifically, we use the Gram-Schmidt orthonormalization process.
Let $v' = v/\norm{v}$ and $y' = y - (y \cdot v')v'$.
  The canonicalizing transformation is then
  \begin{align}
    R_c &= \left[\frac{y'}{\norm{y'}} \times v' \quad \frac{y'}{\norm{y'}} \quad v'\right] \\
    T &= \left[R_c^\top \mid -R_c^\top t \right]
\end{align}
where $T\in \SE(3)$.
We apply $T$ to every camera pose, which results in the target ray having origin $(0, 0, 0)$
and direction $(0, 0, 1)$, and all other ray representations computed from the canonicalized camera poses
will be invariant to similarities.
We show the benefit of such canonicalization in Section~\ref{sec:ablation}.

\subsection{Rendering network}
\label{sec:rendering}
Given the patch embeddings and positional encodings of a target ray, as described in
\cref{sec:posenc,sec:canon}, our rendering network predicts the ray color.

We argue that predicting the target ray color is deeply related to finding correspondences to the target ray in the reference images. 
Take, for example, LFNR~\cite{suhail21_light_field_neural_render}.
Its first stage aggregates features along each epipolar line, 
which is essentially finding correspondences to the target ray.
Since LFNR overfits to a single scene, the model can learn the structure of the scene
and use it to estimate correspondences based only on ray coordinates.

However, the LFNR~\cite{suhail21_light_field_neural_render} approach cannot generalize
to novel scenes, since, given just an epipolar line, it is impossible to know which point corresponds
to a target ray without knowing the structure of the scene.

Our main contribution is to provide visual features for a similar epipolar transformer,
such that the correspondence is solved visually (see \cref{fig:patch} for illustration),
which is advantageous because the visual features
can be extracted from novel scenes in a single forward step starting from small local patches.
Crucially, such features cannot come from a single epipolar line.
It is the combination of visual features from different epipolar lines cast by the same target ray
that allows correspondences to be established.
To learn this combination, we propose to use a transformer. 

Thus, our model consists of three transformers.
The first,  which we call ``Visual Feature Transformer'',
learns visual features by combining information from patches along different reference views.
The second and third are similar to the ones in LFNR~\cite{suhail21_light_field_neural_render},
with the major differences that the positional encodings of rays,
depths and cameras are canonicalized as described in \cref{sec:canon}
and that the final color is predicted by directly blending pixel colors from reference views,
instead of using learned features; both changes greatly improve the generalization performance.

Each transformer follows the ViT~\cite{dosovitskiy2020image} architecture,
which uses residual connections to interleave layer normalization (LN), self-attention (SA),
and multi-layer perceptron (MLP).
Each layer consists of LN \textrightarrow\, SA \textrightarrow\, LN \textrightarrow\, MLP.

\subsubsection{Visual Feature Transformer.}
This stage exchanges visual information between potentially corresponding patches on different
reference images, leading to visual features with multi-view awareness.
The input to this stage is the set of patch linear embeddings and positional encoding vectors
$p_k^m,\, r_k^m,\, d^m,\, c_k$,
indexed by the view $k$ and the $m$-th sampled depth, as described in \cref{sec:posenc}.
We first define the feature concatenation at layer zero (the input) as
\begin{align}
 f_0^{k,m} &= [ p_k^m \;\lVert\; r_k^m \;\lVert\; d^m \;\lVert\; c_k ]. %
\end{align}

This stage is repeated for each depth sample, therefore it operates on sequences of
$K$ views. Formally, it repeats
\begin{align}
  f_1^{m} = T_1\left(\left\{f_0^{k,m} \mid 1 \le k \le K\right\}\right)
\end{align}
for $1 \le m \le M$, where $T_1$ is a transformer written as a set to set map.
This stage takes a $(K, M, C_0)$ tensor of $C_0$-dimensional features of
$K$ views sampled at $M$ depths, 
and returns a $(K, M, C_1)$ tensor of $C_1$-dimensional features.

\subsubsection{Epipolar Aggregator Transformer.}
This stage aggregates information along each epipolar line,
resulting in per reference view features.
The input to this stage is the set $f_1 = \{f_1^m \mid 1 \le m \le M\}$,
concatenated with positional encodings.
We refer to the features corresponding to view $k$ in the set $f_1^m$ as $f_1^{k, m}$.
The transformer is repeated for each view, therefore operating along the sequence of
$M$ epipolar line samples.
Formally, we first compute
\begin{align}
  f_2^{k} = T_2\left(
  \Big\{ r^0 \Big\} \bigcup
  \left\{\left[f_1^{k,m} \;\Big\|\; r_k^m \;\Big\|\; d^m \;\Big\|\; c_k \right]
  \;\Big|\; 1 \le m \le M\right\}\right),
\end{align}
for $1 \le k \le K$, where $r^0$ is a special token to represent the target ray. 
We then apply a learned weighted sum along the $M$ epipolar line samples as follows,
\begin{align}
  \label{eq:alpha}
  \alpha_{k}^{m} &= \frac{\exp \left(W_{1} \left[f_2^{k,0} \;\Big\|\; f_2^{k, m}\right] \right)}
  { \sum\limits_{m'=1}^M \exp \left(W_{1} \left[f_2^{k,0} \;\Big\|\; f_2^{k, m'}\right] \right)}, \\
  f_{2'}^k &= \sum_{m=1}^M \alpha_{k}^{m} f_2^{k, m},
\end{align}
for $1 \le k \le K$,
resulting in a feature vector per view $k$,
where $W_1$ are learnable weights and 
$f_2^{k,0}$ is the output corresponding to the target ray token.

Dimension-wise, this stage takes a $(K, M, C_1)$ tensor 
and returns a $(K, C_2)$ tensor of $C_2$-dimensional features per reference view.

\subsubsection{Reference View Aggregator Transformer.}
This final transformer aggregates the features over reference views
and predicts the color of the target ray.
Its input is the set of per reference view features $f_{2'} = \{ f_{2'}^k \mid 1 \le k \le K \}$,
concatenated with the camera relative positional encoding.
Formally, we compute
\begin{align}
  f_3 = T_3\left(
  \Big\{ r^0 \Big\} \bigcup \left\{\left[f_{2'}^{k} \;\Big\|\; c_k \right]
  \;\Big|\; 1 \le k \le K\right\}\right).
\end{align}
Similarly to the previous stage, we compute the blending weights
\begin{align}
  \label{eq:beta}
  \beta_{k} &= \frac{\exp \left(W_{2} \left[f_3^0 \;\Big\|\; f_3^{k}\right] \right)}
  { \sum\limits_{k'=1}^K \exp \left(W_{2} \left[f_3^{0} \;\Big\|\; f_3^{k}\right] \right)},
\end{align}
which are used in conjunction with the weights from the previous stage
to estimate the color of the target ray by
blending colors along each epipolar line sample at each reference view,
\begin{align}
  \mathfrak{c} = \sum\limits_{k=1}^K \beta_k \left(\sum\limits_{m=1}^M \alpha_k^m \mathfrak{c}_k^m\right),
\end{align}
where $\mathfrak{c}_k^m$ is the pixel color at the $m$-th sample along the epipolar line of view $k$.
Our approach here differs from the last stage of LFNR,
which does the aggregation on feature space using only the weights $\beta_k$,
and linearly projects the resulting feature to predict the color.
We argue that using the input pixel values from reference views instead helps generalization, which we confirm experimentally (see appendix).
This is possible by using the two sets of attention weights $\alpha_k^m$ (\cref{eq:alpha})
and $\beta_k$ (\cref{eq:beta}),
which allow blending colors from all epipolar line samples and all reference views.

\section{Experiments}
\subsection{Implementation Details}
Each of the three transformers in our model consist of 8 blocks each with a feature dimension of $256$.
We select reference views using $K=10$ and $N=20$ (see \cref{sec:patchextraction}).
We use a batch size of $4096$ rays and train for $250k$ iterations with a Adam optimizer
and initial learning rate of $3\cdot 10^{-4}$.
We use a linear learning rate warm-up for $5k$ iterations and cosine decay afterwards. Training our model takes ${\sim}24$ hours on $32$ TPUs.
We report the average PSNR (peak signal-to-noise ratio),
SSIM (structural similarity index measure)
and LPIPS (learned perceptual image patch similarity) for all our experiments.

\begin{figure}[t!]
    \centering
    \includegraphics[width=.82\textwidth]{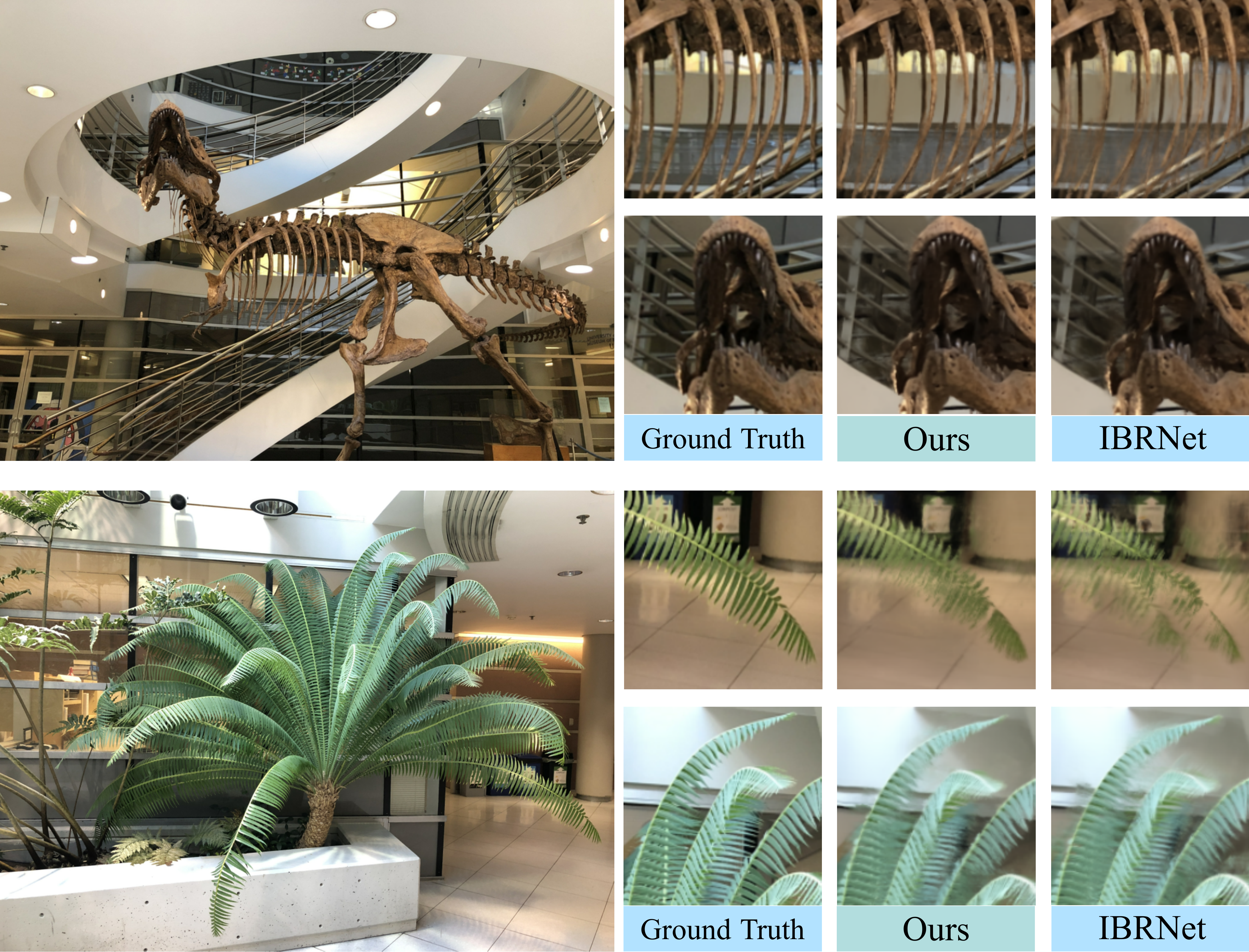}
    \caption{\textbf{Qualitative results on RFF (setting 1).}
      We show our method and the baseline on the \emph{T-Rex} and \emph{Fern} scenes from the real forward-facing dataset.
      Compared with IBRNet~\cite{wang2021ibrnet},
      our method produce sharper details and less blurring at boundaries.
      For example, the top row in the \emph{Fern} scene shows that the baseline methods
      either fail to reconstruct the leaves or produce inconsistent shapes.
      Our method is able to retain the shape boundaries accurately
      along with majority of the texture details.}
    \label{fig:rff_qualitative}
  \end{figure}
 
\begin{table}[t!]
\centering
\setlength{\tabcolsep}{2.9pt}  
\begin{tabular}{@{}lccccccccccc}
\toprule
\toprule
Method                           & \multicolumn{3}{c}{Real Forward-Facing} &           & \multicolumn{3}{c}{Shiny-6} &  & \multicolumn{3}{c}{Blender}                                              \\ \cmidrule(lr){2-4} \cmidrule(lr){6-8} \cmidrule(l){10-12} 
                                 & PSNR                                    & SSIM      & LPIPS                       &  & PSNR      & SSIM      & LPIPS     &  & PSNR      & SSIM      & LPIPS     \\ \midrule
LLFF~\cite{mildenhall2019local}  & 24.13                                   & 0.798     & 0.212                       &  & -         & -         & -         &  & 24.88     & 0.911     & 0.114     \\
IBRNet~\cite{wang2021ibrnet}     & 25.13                                   & 0.817     & 0.205                       &  & \bgl23.60 & \bgl0.785 & \bgl0.180 &  & 25.49     & 0.916     & 0.100     \\
GeoNeRF~\cite{johari2021geonerf} & \bgl25.44                               & \bgl0.839 & \bgl0.180                   &  & -         & -         & -         &  & \bgd28.33 & \bgl0.938 & \bgd0.087 \\ \midrule
IBRNet*                          & 24.33                                   & 0.801     & 0.213                       &  & 23.37     & 0.784     & 0.181     &  & 21.32     & 0.888     & 0.131     \\
Ours                             & \bgd25.72                               & \bgd0.880 & \bgd0.175                   &  & \bgd24.12 & \bgd0.860 & \bgd0.170 &  & \bgl26.48 & \bgd0.944 & \bgl0.091 \\ \bottomrule \bottomrule
\end{tabular}
\caption{{\bf Results for setting 1.}
  Our model outperforms the baselines even when training with strictly less data.
  IBRNet uses three datasets that are not part of our training set,
  while  GeoNeRF uses one extra dataset and also leverages input depth maps during training.
  IBRNet* was trained using the same training set as our method;
  in this fair comparison, our advantage in accuracy widens.
}
\label{table:all_res}
\end{table}
\subsection{Results}
There is no standard training and evaluation procedure for generalizable neural rendering.
IBRNet~\cite{wang2021ibrnet} trains on the LLFF dataset~\cite{mildenhall2019local},
renderings of Google scanned objects~\cite{googlescanned},
Spaces dataset~\cite{flynn2019deepview},
RealEstate10K dataset~\cite{zhou2018stereo}
and on their own scenes.
They evaluate on the real forward-facing (RFF) dataset~\cite{mildenhall2020nerf}, which comprises held-out LLFF scenes, Blender (consisting of $360^\circ$ scenes)~\cite{mildenhall2020nerf},
and Diffuse Synthetic $360^\circ$~\cite{sitzmann2019deepvoxels}.
Contrastingly, MVSNeRF~\cite{chen2021mvsnerf} trains on DTU~\cite{jensen2014large}
and tests on held out DTU scenes,
real forward-facing dataset (RFF)~\cite{mildenhall2020nerf},
and Blender \cite{mildenhall2020nerf}.
Various other works \cite{johari2021geonerf,liu2021neural,trevithick2021grf} have explored different experimental setups.
In this work, in an attempt to fairly evaluate against prior works, we use two experimental settings.

\subsubsection{Setting $1$.}
\begin{figure}[t!]
    \centering
    \includegraphics[width=.82\textwidth]{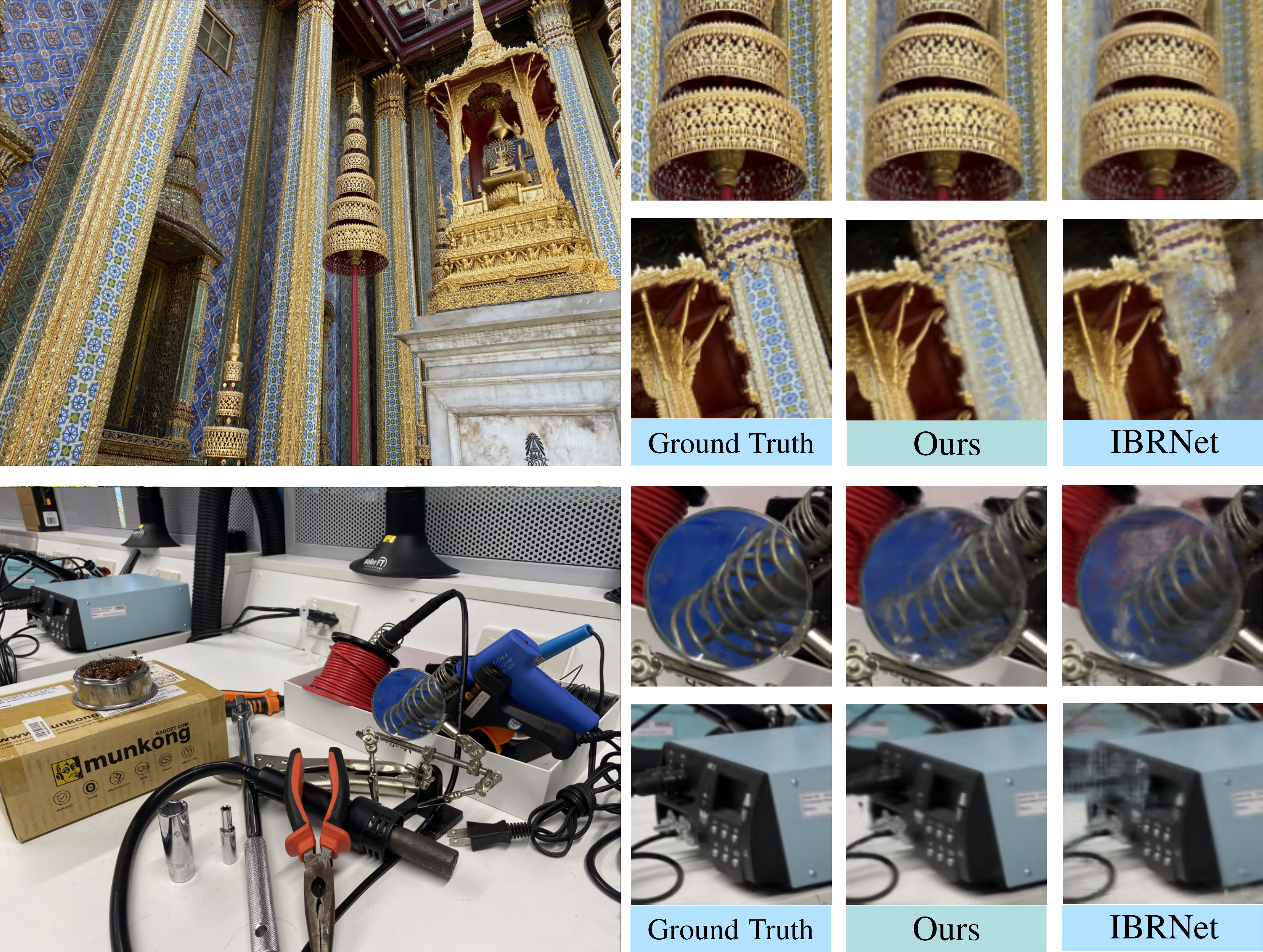}
    \caption{\textbf{Qualitative results on Shiny~\cite{wizadwongsa2021nex} (setting 1).}
      While still consisting of forward facing scenes,
      Shiny scenes have scale and view density that differ from the usual in setting 1, which makes it more challenging than LLFF.
      IBRNet~\cite{wizadwongsa2021nex}
      produces noticeable artifacts that are not present in our method's renderings.
    }
    \label{fig:shiny_qualitative}
\end{figure}

In the first setting, we train on a strict subset of the IBRNet training set,
comprised of $37$ LLFF scenes and $131$ IBRNet collected scenes
(amounting to $11\%$ of the training set used by IBRNet).
We then evaluate on the real forward-facing, Shiny~\cite{wizadwongsa2021nex} and Blender datasets.
On Shiny, we compute the results for IBRNet using their publicly available pretrained weights.
\Cref{table:all_res} reports quantitative
while \cref{fig:rff_qualitative,fig:shiny_qualitative} show qualitative results.
IBRNet and GeoNeRF (a concurrent work) use a larger training set than ours,
and GeoNeRF uses depth maps during training,
but our method shows the best performance in most metrics regardless.
Additionally, IBRNet is trained on \ang{360} scenes whereas
our method is trained only on forward-facing scenes.
Nonetheless, our model achieves superior performance on Blender as compared to IBRNet.

\subsubsection{Setting $2$.}
Here, we train our model on DTU, following the MVSNeRF~\cite{Chen_2021_ICCV} procedure,
and evaluate on the held-out DTU scenes and the Blender dataset.
For training on DTU, we follow the same split as PixelNeRF~\cite{Yu_2021_CVPR} and MVSNeRF.
We partition the dataset into $88$ scenes for training and $16$ scenes for testing,
each containing images of resolution $512 \times 640$.
\Cref{table:dtu} shows quantitative results.
MVSNeRF is trained with 3 reference views while our method performs best with 10.
We evaluated MVSNeRF with 10 views, which did not improve their performance;
\cref{table:dtu}, thus, compares the best number of views for each model.
Our model consistently outperforms across all three metrics.

\subsubsection{Ablation.}
\label{sec:ablation}
To investigate the effectiveness of our contributions, we perform various ablations experiments. We train the model on $504 \times 378$ resolution images of LLFF and IBRNet scenes and test on the real forward-facing dataset at the same resolution.
We start with a ``base model''
that does not use the visual feature transformer or the coordinate canonicalization. We then incrementally add components of our proposed approach. Table~\ref{table:ablation} reports the ablation results. We observe that the ``base'' model generalized poorly to unseen scenes.
Incorporating the visual feature transformer improves the performance significantly. For the canonicalization ablation, we split the component into two,
(1) ray canonicalization, where the light field ray representation is computed independent of the frame of reference, and
(2) coordinate canonicalization where the $3D$ samples along the target ray are canonicalized.
We observe that both forms of canonicalization help improve the accuracy.

\begin{table}[ht]
  \centering
\begin{tabular}{@{}lccccccc}
\toprule
\toprule
  \multirow{2}{*}{Method}       & \multicolumn{3}{c}{DTU} & \multicolumn{1}{l}{} & \multicolumn{3}{c}{Blender}                          \\
  \cmidrule(l){2-4} \cmidrule(l){6-8} 
                                & PSNR                    & SSIM                 & LPIPS      &  & PSNR       & SSIM       & LPIPS      \\
  \midrule
 PixelNeRF~\cite{Yu_2021_CVPR}  & 19.31                   & 0.789                & 0.671      &  & 7.39       & 0.658      & 0.411      \\
 IBRNet~\cite{wang2021ibrnet}  & 26.04                   & 0.917                & 0.190       &  & 22.44      & 0.874      & 0.195  \\
 MVSNeRF~\cite{chen2021mvsnerf} & \bgl26.63               & \bgl0.931            & \bgl 0.168 &  & \bgl23.62  & \bgl0.897  & \bgl0.176  \\
  Ours                          & \bgd 28.50              & \bgd0.932            & \bgd 0.167 &  & \bgd 24.10 & \bgd 0.933 & \bgd 0.097 \\
  \bottomrule\bottomrule
\end{tabular}
\caption{{\bf Results for setting 2.}
  All models are trained on DTU and evaluated on either the DTU held-out set or Blender.
  Our approach outperforms the baselines.
}
\label{table:dtu}
\end{table}

\begin{table}[]
\small
\centering
\begin{tabular}{@{}cccccc@{}}
\toprule\toprule
  Visual   & Ray&  Coordinate & \multirow{2}{*}{PSNR}  & \multirow{2}{*}{SSIM}  & \multirow{2}{*}{LPIPS} \\
  Transformer  & Canonicalization & Canonicalization &   &   &  \\  
  \midrule
\xmark & \xmark & \xmark &  22.62     &  0.763     &     0.313  \\
\cmark & \xmark & \xmark & 25.42 & 0.879 & 0.154 \\
\cmark & \cmark &  \xmark & 25.86 & 0.885 & 0.142 \\
  \cmark & \cmark & \cmark & 26.42 & 0.896 & 0.129 \\
  \bottomrule\bottomrule
\end{tabular}
\caption{{\bf Ablations.}
  Ablation study for model trained on LLFF and IBRNet scenes
  and tested on RFF with a resolution of $504 \times 378$.
  Results show that our main contributions -- the visual feature transformer and
  the canonicalized positional encoding -- lead to superior generalization performance.
}
\label{table:ablation}
\end{table}

\section{Limitations}

One limitation of our model is that since it operates on small local patches to aid generalization,
it relies on a large number of views to produce meaningful features.
In the comparison against MVSNeRF~\cite{Chen_2021_ICCV} in \cref{table:dtu},
while our method is more accurate by significant margins,
it also requires 10 reference views while MVSNeRF only uses 3.
Rendering novel scenes with our approach is fast since it 
consists only of forward steps, but training is slow, comparable with
LFNR~\cite{suhail21_light_field_neural_render}.
The appendix shows a quantitative timing evaluation.

\section{Conclusion}
This paper introduced a method to generate novel views from unseen scenes
that predicts the color of an arbitrary ray directly from a collection of small local patches
sampled from reference views according to epipolar constraints.
Our model departs from the common combination of using deep visual features
and NeRF-like volume rendering for this task.
We introduced a three-stage transformer architecture,
coupled with canonicalized positional encodings,
which operates on local patches -- all these properties aid in generalizing to unseen scenes.
This is demonstrated by our outperforming of the current state-of-the-art while using only 11\%
of the amount of training data.

We include more details and results in the appendix,
including more ablation experiments, timing evaluation,
other combinations of train and evaluation sets,
and more qualitative results.

\clearpage
\bibliographystyle{splncs04}
\bibliography{egbib}
\end{document}


\pagestyle{headings}
\mainmatter
\def\ECCVSubNumber{2290}  %

\title{Generalizable Patch-Based Neural Rendering \\ Supplementary Material} %

\titlerunning{Generalizable Patch-Based Neural Rendering}
\author{Mohammed Suhail\inst{1} \and
Carlos Esteves\inst{4} \and
Leonid Sigal\inst{1,2,3}\and 
Ameesh Makadia\inst{4}
}
\authorrunning{M. Suhail et al.}
\institute{University of British Columbia \email{\{suhail33,lsigal\}@cs.ubc.ca} \and Vector Institute for AI \and
Canada CIFAR AI Chair \and
Google
\email{\{machc,makadia\}@google.com}
}

\maketitle

\section{Additional Experiments and Results}

\subsection{Fine-tuning}
While our model is focused at generalizing to unseen scenes, for a thorough comparison against previous methods, we follow the protocol from IBRNet~\cite{wang2021ibrnet} and fine-tune our model (for setting 1) on each of the RFF test scenes for 10k iterations.
We report the average metrics across all scenes in Table~\ref{table:fine-tune}.
\begin{table}[]
\centering
\begin{tabular}{@{}lccc}
\toprule
\toprule
Method                           & PSNR       & SSIM       & LPIPS      \\ \midrule
IBRNet~\cite{wang2021ibrnet}     & \bgl 26.73 & 0.851      & 0.175      \\
GeoNeRF~\cite{johari2021geonerf} & 26.58      & 0.\bgl 856 & \bgl 0.162 \\
Ours                             & \bgd 27.66 & \bgd 0.924 & \bgd 0.138 \\ \bottomrule \bottomrule
\end{tabular}
\caption{Fine-tuning results on the RFF dataset, setting 1.
  Our approach not only improves over the baselines on unseen scenes with no re-training
  as shown in the main paper,
  but also when allowed to fine-tune for a few iterations on new scenes.}%
\label{table:fine-tune}
\end{table}

\subsection{Number of Reference Views}
We train our model with $3,5,7, 10$ and $12$ reference images
to investigate the effect of number of reference views available to view synthesis.
The models are trained on forward-facing scenes from LLFF~\cite{mildenhall2019local} and IBRNet~\cite{wang2021ibrnet}. We summarize the average performance of each variant on the real-forward-facing dataset in \cref{table:num_ref}.
Our model benefits from having access to a large number (up to 10) of reference images.

\begin{table}[]
\centering
\begin{tabular}{@{}cccc@{}}
\toprule \toprule
\multirow{2}{*}{\begin{tabular}[c]{@{}c@{}}Number of\\ Reference Views\end{tabular}} & \multicolumn{3}{c}{Real-Forward-Facing} \\ \cmidrule(l){2-4} 
                                                                                     & PSNR        & SSIM        & LPIPS       \\ \midrule
3                                                                                    & 22.36       & 0.800       & 0.286       \\
5                                                                                    & 24.33       & 0.850       & 0.216       \\
7                                                                                    & 25.05       & 0.860       & 0.195       \\
10                                                                                   & 25.72       & 0.880       & 0.175       \\
12                                                                                   & 25.69       & 0.881       & 0.178       \\ \bottomrule \bottomrule
\end{tabular}
\caption{Effect of varying the number of reference view on view-synthesis.}
\label{table:num_ref}
\end{table}

\subsection{RGB Prediction}
To synthesize a novel view,
our model predicts the weights of a linear combination of reference image pixel colors.
A common alternative is to combine learned visual features, followed by a learned mapping to color values~\cite{suhail21_light_field_neural_render}.
To substantiate our argument of better generalization from combining colors instead of features,
we train a variant of our model where the output color is predicted from the aggregated features as
\begin{align}
  \mathfrak{c} = \texttt{MLP}\left(\sum\limits_{k=1}^K \beta_k f_{3}^{k}\right).
\end{align}
We evaluate this approach in setting 1 and show that
our model, which combines pixels, is indeed superior.
\Cref{table:rgb} shows the results.
\begin{table}[]
\centering
\begin{tabular}{@{}cccc@{}}
\toprule \toprule
\multirow{2}{*}{\begin{tabular}[c]{@{}c@{}}Interpolation\\ Method\end{tabular}} & \multicolumn{3}{c}{Real-Forward-Facing} \\ \cmidrule(l){2-4} 
                                                                                & PSNR         & SSIM       & LPIPS       \\ \midrule
Features                                                                         & 25.08        & 0.86       & 0.199       \\
Colors (ours)                                                                             & 25.72        & 0.88       & 0.175       \\ \bottomrule \bottomrule
\end{tabular}
\caption{Comparison of average performance when using feature versus color interpolation for view-synthesis.
Results show that combining the colors of reference views generalizes better than combining visual features.}
\label{table:rgb}
\end{table}

\subsection{DTU to RFF Generalization}
We present results on generalization to scenes in the real-forward-facing (RFF) dataset for a model trained only on DTU in \cref{table:dtu2rff}. While MVSNeRF~\cite{Chen_2021_ICCV} has a better PSNR and LPIPS performance our method achieves better SSIM scores on average across all scenes in RFF.

\begin{table}[]
\centering
\begin{tabular}{@{}lccc}
\toprule \toprule
\multirow{2}{*}{Method~} & \multicolumn{3}{c}{Real-Forward-Facing} \\ \cmidrule(l){2-4} 
                         & PSNR  & SSIM  & LPIPS                   \\ \midrule
MVSNeRF                  & 21.93 & 0.795 & 0.252                   \\
Ours                     & 20.69 & 0.808 & 0.281                   \\ \bottomrule \bottomrule
\end{tabular}
\caption{Generalization results on RFF for a model trained on DTU.}
\label{table:dtu2rff}
\end{table}

\subsection{Timing Statistics}
On average, our model trains at around $3.2$ steps per second on $32$ TPUs with a batch size of $4096$.
A prior transformer-based neural rendering work, LFNR~\cite{suhail21_light_field_neural_render}, is slightly faster at $4.2$ steps per second on the same hardware with the same batch size.
Rendering one image with our model takes around $15$ seconds whereas LFNR takes around $10$ seconds on the same hardware.
While LFNR takes around $16$ hours to train, it can only be trained on single scene.
Thus, to render the  novel views for the $8$ scenes in the RFF dataset,
it would take at least $128$ hours of training plus the inference time.
Our model just trains for approximately $24$ hours
and can be used for inference directly on all the scenes,
albeit with a small drop in rendering quality. 

\section{Quatlitative Results}
\subsection{DTU Comparison}
\begin{figure}[htp]
    \centering
    \includegraphics[width=\textwidth]{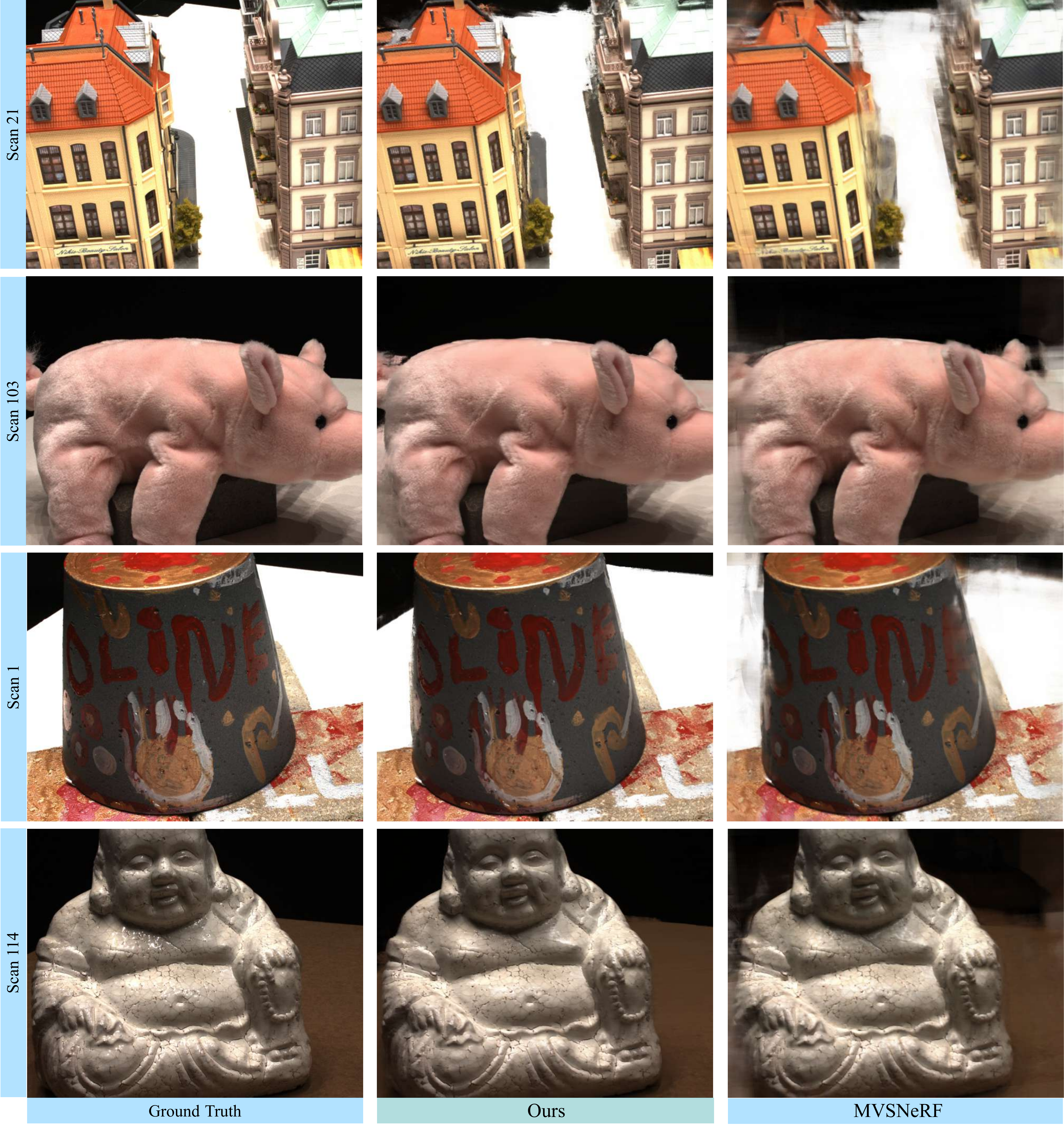}
    \caption{\textbf{Qualitative Results on DTU.}}
    \label{fig:dtu_qualitative}
\end{figure}

We compare renderings on the DTU test set against MSVNeRF in \cref{fig:dtu_qualitative}.
Compared to MVSNeRF, our model produces renderings with sharper boundaries and textures.

\bibliographystyle{splncs04}
\bibliography{egbib}